\documentclass[11pt,a4paper]{article}
\usepackage[hyperref]{emnlp2018}
\usepackage{CJKutf8}
\usepackage[dvipdfmx]{graphicx}
\usepackage{graphicx}
\usepackage{times}
\usepackage{latexsym}
\usepackage{ascmac}
\usepackage{amsmath}
\usepackage{relsize}
\usepackage[T1]{fontenc}
\usepackage{fancyvrb}
\usepackage{listings}
\usepackage{setspace}

\usepackage{url}

\aclfinalcopy

\hypersetup{
    colorlinks=true,
    linkcolor=black,
    citecolor=black,
    filecolor=black,
    urlcolor=black,
}

\DeclareMathOperator{\Encode}{Encode}
\DeclareMathOperator{\Decode}{Decode}
\DeclareMathOperator{\Normalize}{Normalize}

\let\olditemize\itemize
\renewcommand{\itemize}{
\olditemize
\setlength{\itemsep}{1.2pt}
\setlength{\parskip}{0pt}
\setlength{\parsep}{0pt}
}

\title{SentencePiece: A simple and language independent subword
 tokenizer and detokenizer for Neural Text Processing}

\author{Taku Kudo \hspace{4em} John Richardson \\
Google, Inc.\\
  {\tt \{taku,johnri\}@google.com} \\
}

\date{}

\begin{document}
\maketitle

\begin{abstract}
This paper describes SentencePiece, a language-independent subword
tokenizer and detokenizer designed for Neural-based text processing,
including Neural Machine Translation. It provides open-source C++ and
Python implementations for subword units.  While existing subword
segmentation tools assume that the input is pre-tokenized into word
sequences, SentencePiece can train subword models directly from raw
sentences, which allows us to make a purely end-to-end and language
independent system. We perform a validation experiment of NMT on
English-Japanese machine translation, and find that it is possible to
achieve comparable accuracy to direct subword training from raw
sentences. We also compare the performance of subword training and
segmentation with various configurations.  SentencePiece is available
under the Apache 2 license at 
\url{https://github.com/google/sentencepiece}.
\end{abstract}

\section{Introduction}
Deep neural networks are demonstrating a large impact on Natural Language Processing.
Neural machine translation (NMT) \cite{bahdanau2014neural,luong2015effective,wu2016google,ashish2017google}
has especially gained increasing popularity, as it can leverage neural
networks to directly perform translations with a simple end-to-end
architecture. NMT has shown remarkable results in several shared tasks
\cite{denkowski2017stronger,nakazawa2017overview}, and its effective
approach has had a strong influence on other related NLP tasks such as
dialog generation \cite{vinyals20152} and automatic summarization
\cite{rush2015neural}.

Although NMT can potentially perform end-to-end translation, many NMT
systems are still relying on language-dependent pre- and postprocessors,
which have been used in traditional statistical machine translation
(SMT) systems. Moses\footnote{\url{http://www.statmt.org/moses/}}, a
de-facto standard toolkit for SMT, implements a reasonably useful pre- and
postprocessor. However, it is built upon hand-crafted and language
dependent rules whose effectiveness for NMT has not been proven. In
addition, these tools are mainly designed for European languages where
words are segmented with whitespaces. To train NMT systems for
non-segmented languages such as Chinese, Korean and Japanese, we need to
run word segmenters independently.  Such language-dependent processing
also makes it hard to train multilingual NMT models
\cite{johnson2016google}, as we have to carefully manage the
configurations of pre- and postprocessors per language, while the
internal deep neural architectures are language-independent.

As NMT approaches are standardized and moving forward to more
language-agnostic architectures, it is becoming more important for the NLP
community to develop a simple, efficient, reproducible and language
independent pre- and postprocessor that can easily be integrated into
Neural Network-based NLP systems, including NMT.

In this demo paper, we describe SentencePiece, a simple and language
independent text tokenizer and detokenizer mainly for Neural
Network-based text generation systems where the size of vocabulary is
predetermined prior to the Neural model training. SentencePiece
implements two subword segmentation algorithms, byte-pair-encoding (BPE)
\cite{sennrichneural} and unigram language model \cite{kudo2018}, with
the extension of direct training from raw sentences.  SentencePiece
enables building a purely end-to-end system that does not depend on any
language-specific processing.

\section{System Overview}
SentencePiece comprises four main components: {\bf Normalizer}, {\bf
Trainer}, {\bf Encoder}, and {\bf Decoder}. Normalizer is a module to
normalize semantically-equivalent Unicode characters into canonical
forms.  Trainer trains the subword segmentation model from the
normalized corpus. We specify a type of subword model as the parameter
of Trainer. Encoder internally executes Normalizer to normalize the
input text and tokenizes it into a subword sequence with the subword
model trained by Trainer. Decoder converts the subword sequence into the
normalized text.

The roles of Encoder and Decoder correspond to preprocessing
(tokenization) and postprocessing (detokenization)
respectively. However, we call them encoding and decoding as
SentencePiece manages the vocabulary to id mapping and can directly
convert the text into an id sequence and vice versa. Direct encoding and
decoding to/from id sequences are useful for most of NMT systems
as their input and output are id sequences.

Figure \ref{fig1:command} presents end-to-end example of SentencePiece
training (\verb+spm_train+), encoding (\verb+spm_encode+), and decoding
(\verb+spm_decode+). We can see that the input text is
reversibly converted through \verb+spm_encode+ and \verb+spm_decode+.

\begin{figure}[t]
\begin{lstlisting}[language=sh, caption=Commandline usage of SentencePiece, label=fig1:command
 language={python},
 lineskip=-0.5ex,%
 basicstyle={\normalsize},%
 identifierstyle={\normalsize},%
 commentstyle={\normalsize\itshape},%
 keywordstyle={\normalsize\bfseries},%
 ndkeywordstyle={\normalsize},%
 stringstyle={\normalsize\ttfamily}]
% spm_train --input=data/input.txt
 --model_prefix=spm --vocab_size=1000

% echo "Hello world." | spm_encode --model=spm.model
_He ll o _world .

% echo "Hello world." | spm_encode
 --model=spm.model --output_format=id
151 88 21 887 6

% echo "_He ll o _world ." | spm_decode --model=spm.model
Hello world.

% echo "151 88 21 887 6" | spm_decode --model=spm.model
       --input_format=id
Hello world.
 \end{lstlisting}
 \vspace*{-4mm}
\end{figure}

\section{Library Design}
This section describes the design and implementation details of
SentencePiece with command line and code snippets.

\subsection{Lossless Tokenization}
The following raw and tokenized sentences are an
example of language-dependent preprocessing.
 \begin{itemize}
\item {\bf Raw text:}\,\,   Hello world.
\item {\bf Tokenized:}  [Hello] [world] [.]
\end{itemize}
One observation is that the raw text and tokenized sequence are not
reversibly convertible.  The information that no space exists between
``world'' and ``.'' is not kept in the tokenized sequence.
Detokenization, a process to restore the original raw input from the
tokenized sequence, has to be language-dependent due to these
irreversible operations. For example, while the detokenizer usually puts
whitespaces between the primitive tokens in most European languages, no
spaces are required in Japanese and Chinese.
 \begin{itemize}
  \setlength{\leftskip}{-1em}
  \item {\bf Raw text:}  \begin{CJK}{UTF8}{ipxm}[こんにちは世界。]\end{CJK} ({\it Hello world.})
  \item {\bf Tokenized:} \begin{CJK}{UTF8}{ipxm}[こんにちは] [世界] [。]\end{CJK}
 \end{itemize}
Such language specific processing has usually been implemented in
manually crafted rules, which are expensive to write and
maintain.

SentencePiece implements the Decoder as an inverse operation of Encoder,
i.e.,
\begin{eqnarray*}
 \Decode(\Encode(\Normalize(text))) = \\
   \Normalize(text).
\end{eqnarray*}
We call this design {\bf lossless tokenization}, in which all the
information to reproduce the normalized text is preserved in the
encoder's output. The basic idea of lossless tokenization is
to treat the input text just as a sequence of Unicode characters. Even
whitespace is handled as a normal symbol. For the sake of clarity,
SentencePiece first escapes the whitespace with a meta symbol
\textunderscore\,\,(U+2581), and tokenizes the input into an
arbitrary subword sequence, for example:
\begin{itemize}
 \item {\bf Raw text:}\,\,\,\,Hello\textunderscore world.
 \item {\bf Tokenized:}   [Hello] [\textunderscore wor] [ld] [.]
\end{itemize}
As the whitespace is preserved in the tokenized text, we can
detokenize the tokens without any ambiguities with the following Python code.

{\small
\begin{Verbatim}
detok = ''.join(tokens).replace('_', ' ')
\end{Verbatim}
}

It should be noted that subword-nmt\footnote{\url{https://github.com/rsennrich/subword-nmt}}
adopts a different representation for subword units. It focuses on how the word
is segmented into subwords and uses \verb+@@+ as an intra-word boundary marker.
\begin{itemize}
 \item {\bf Tokenized:} [Hello] [wor] [\verb+@@+ld] [\verb+@@+.]
\end{itemize}
This representation can not always perform lossless tokenization, as
an ambiguity remains in the treatment of whitespaces. More specifically,
it is not possible to encode consecutive whitespaces with this
representation.

\subsection{Efficient subword training and segmentation}
Existing subword segmentation tools train subword models from
pre-tokenized sentences. Such pre-tokenization was introduced for an
efficient subword training \cite{sennrichneural}. However, we can not
always assume that pre-tokenization is available, especially for
non-segmented languages.  In addition, pre-tokenization makes it
difficult to perform lossless tokenization.

SentencePiece employs several speed-up techniques both for training and
segmentation to make lossless tokenization with a large amount of
raw data. For example, given an input sentence (or word) of length $N$,
BPE segmentation requires $O(N^2)$ computational cost when we naively
scan the pair of symbols in every iteration.  SentencePiece adopts
an $O(N\log(N))$ algorithm in which the merged symbols are managed by a
binary heap (priority queue). In addition, the training and segmentation
complexities of unigram language models are linear to the size of input
data.


\subsection{Vocabulary id management}
SentencePiece manages the vocabulary to id mapping to directly convert
the input text into an id sequence and vice versa. The size of
vocabulary is specified with the \verb+--vocab_size=<size>+ flag of
\verb+spm_train+. While subword-nmt specifies the number of merge
operations, SentencePiece specifies the final size of vocabulary, as the
number of merge operations is a BPE specific parameter and can not be
applicable to other segmentation algorithms, e.g., unigram language model
\cite{kudo2018}.

SentencePiece reserves vocabulary ids for special meta symbols, e.g.,
unknown symbol
(\verb+<unk>+), BOS (\verb+<s>+), EOS (\verb+</s>+) and padding
(\verb+<pad>+).
Their actual ids are configured with command line flags.  We can
also define custom meta symbols to encode contextual information as
virtual tokens. Examples include the language-indicators,
\verb+<2ja>+ and \verb+<2de>+, for multilingual models
\cite{johnson2016google}.

\subsection{Customizable character normalization}
Character normalization is an important preprocessing step for handling
real world text, which consists of semantically-equivalent Unicode
characters.  For example, Japanese fullwidth Latin characters can be
normalized into ASCII Latin characters. Lowercasing is also an
effective normalization, depending on the application.

Character normalization has usually been implemented as hand-crafted
rules. Recently, Unicode standard Normalization Forms, e.g., NFC and
NFKC, have been widely used in many NLP applications because of their
better reproducibility and strong support as Unicode standard.

By default, SentencePiece normalizes the input text with the Unicode
NFKC normalization. The normalization rules are specified with the
\verb+--normalization_rule_name=nfkc+ flag of \verb+spm_train+.  The
normalization in Sentencepiece is implemented with string-to-string
mapping and leftmost longest matching. The normalization rules are
compiled into a finite state transducer (Aho-Corasick automaton) to
perform an efficient normalization\footnote{ The original NFKC
normalization requires CCC (Canonical Combining Class) reordering, which
is hard to model in a finite state transducer.  SentencePiece does not
handle the full CCC reordering and only implements a subset of NFKC
normalization.}.

SentencePiece supports custom normalization rules defined as a TSV
file.  Figure \ref{fig1:tsv} shows an example TSV file.
\begin{figure}[t]
\begin{lstlisting}[language=C, caption=Custom normalization rule in TSV, label=fig1:tsv]
U+41 U+302 U+300 <tab> U+1EA6
U+41 U+302 U+301 <tab> U+1EA4
 ...
\end{lstlisting}
\vspace{-6mm}
\end{figure}
In this example, the Unicode sequence [U+41 U+302 U+300] is converted
into U+1EA6\footnote{Note that tabs are used as the delimiter for source
and target sequence and spaces are used as the delimiter for individual
characters.}.  When there are ambiguities in the conversion, the longest
rule is applied. User defined TSV files are specified with the
\verb+--normalization_rule_tsv=<file>+ flag of \verb+spm_train+.
Task-specific rules can be defined by extending the default NFKC rules
provided as a TSV file in SentencePiece package.

\subsection{Self-contained models}
Recently, many researchers have provided pre-trained NMT models for
better reproduciblity of their experimental results. However, it is not
always stated how the data was preprocessed.  \cite{post2018} reported
that subtle differences in preprocessing schemes can widely change BLEU
scores. Even using the Moses toolkit, it is not guaranteed to reproduce
the same settings unless the configurations of Moses (e.g., version and
command line flags) are clearly specified.  Strictly speaking, NFKC
normalization may yield different results depending on the Unicode
version.

Ideally, all the rules and parameters for preprocessing must be embedded
into the model file in a self-contained manner so that we can reproduce
the same experimental setting as long as we are using the same model file.

The SentencePiece model is designed to be purely self-contained. The
model file includes not only the vocabulary and segmentation parameters,
but also the pre-compiled finite state transducer for character
normalization.  The behavior of SentencePiece is determined only by the
model file and has no external dependencies.  This design guarantees a
perfect reproducibility as well as allowing to distribute the
SentencePiece model file as part of an NMT model. In addition, the
developers of SentencePiece can refine the (default) normalization rules
without having to worry about breaking existing preprocessing behaviors.

The SentencePiece model is stored as a binary wire format
Protocol buffer\footnote{\url{https://developers.google.com/}\\\hspace*{30mm}\url{protocol-buffers/}},
a platform neutral and extensible mechanism for serializing structured
data. Protocol buffers help to safely serialize structured data while
keeping backward compatibility as well as extensibility.

\subsection{Library API for on-the-fly processing}
Text preprocessing is usually considered as offline processing.
Prior to the main NMT training, raw input is preprocessed and converted
into an id sequence with a standalone preprocessor.

Such off-line preprocessing has two problems. First, standalone tools
are not directly integrated into the user-facing NMT applications which
need to preprocess user input on-the-fly.  Second, off-line
preprocessing makes it hard to employ sub-sentence level data
augmentation and noise injection, which aim at improving the accuracy
and robustness of the NMT models. There are several studies to inject
noise to input sentences by randomly changing the internal
representation of sentences.  \cite{kudo2018} proposes a subword
regularization that randomly changes the subword segmentation during NMT
training. \cite{guillaume2017,mikel2017} independently proposed a denoising
autoencoder in the context of sequence-to-sequence learning, where they
randomly alter the word order of the input sentence and the model is
trained to reconstruct the original sentence.  It is hard to emulate
this dynamic sampling and noise injection only with the
off-line processing.

\begin{figure}[t]
 \begin{lstlisting}[language=C++, caption=C++ API usage {\small (The same
  as Figure 1.)}, label=fig1:c++]
#include <sentencepiece_processor.h>
#include <sentencepiece_trainer.h>

SentencePieceTrainer::Train(
  "--input=input.txt "
  "--model_prefix=spm "
  "--vocab_size=1000");

SentencePieceProcessor sp;
sp.Load("spm.model");

std::vector<std::string> pieces;
sp.Encode("Hello world.", &pieces);

std::vector<int> ids;
sp.Encode("Hello world.", &ids);

std::string text;
sp.Decode({151, 88, 21, 887, 6}, &text);
 \end{lstlisting}
 \vspace*{-3mm}
\end{figure}

\begin{figure}[t]
\begin{lstlisting}[language=Python, caption=Python API usage {\small (The same
  as Figure 1.)}, label=fig1:python]
import sentencepiece as spm

params = ('--input=input.txt '
         '--model_prefix=spm '
         '--vocab_size=1000')
spm.SentencePieceTrainer.Train(params)

sp = spm.SentencePieceProcessor()
sp.Load('spm.model')

print(sp.EncodeAsPieces('Hello world.'))
print(sp.EncodeAsIds('Hello world.'))
print(sp.DecodeIds([151, 88, 21, 887, 6]))
\end{lstlisting}
 \vspace*{-5mm}
\end{figure}

\begin{figure}[t]
 \begin{lstlisting}[language=Python, label=fig1:tf, caption=TensorFlow
  API usage]
import tensorflow as tf
import tf_sentencepiece as tfs

model = tf.gfile.GFile('spm.model', 'rb').read()

input_text = tf.placeholder(tf.string, [None])
ids, lens = tfs.encode(input_text, model_proto=model, out_type=tf.int32)
output_text = tfs.decode(ids, lens, model_proto=model)

with tf.Session() as sess:
  text = ['Hello world.', 'New York']
  ids_, lens_, output_text_ = sess.run([ids, lens, output_text], feed_dict={input_text:text})
 \end{lstlisting}
 \vspace*{-3mm}
\begin{spacing}{0.5}
{\footnotesize The SentencePiece model (model proto) is an attribute of
the TensorFlow operation and embedded into the TensorFlow graph so the
 model and graph become purely self-contained.}
\end{spacing}
  \vspace*{4mm}
\end{figure}

SentencePiece not only provides a standalone command line tool for
off-line preprocessing but supports a C++, Python and Tensorflow library
API for on-the-fly processing, which can easily be integrated into
existing NMT frameworks. Figures \ref{fig1:c++}, \ref{fig1:python} and
\ref{fig1:tf} show example usages of the C++, Python and TensorFlow
API\footnote{As the Python and TensorFlow wrappers call the native C++
API, there is no performance drop in their interfaces.}.  Figure
\ref{fig1:python2} presents example Python code for subword
regularization where one subword sequence is sampled according to the
unigram language model.  We can find that the text ``New York'' is
tokenized differently on each \verb+SampleEncodeAsPieces+ call. Please see
\cite{kudo2018} for the details on subword regularization and its
sampling hyperparameters.

\begin{figure}[t]
\begin{lstlisting}[language=Python, caption=Subword sampling with Python
 API, label=fig1:python2,lineskip=-0.3ex]
>>> sp.Load('spm.model')
>>> for n in range(5):
...   sp.SampleEncodeAsPieces('New York', -1, 0.1)
['_', 'N', 'e', 'w', '_York']
['_', 'New', '_York']
['_', 'New', '_Y', 'o', 'r', 'k']
['_', 'New', '_York']
['_', 'New', '_York']
\end{lstlisting}
\vspace*{-5mm}
\end{figure}

\section{Experiments}
\subsection{Comparison of different preprocessing}
We validated the performance of the different preprocessing on
English-Japanese translation of Wikipedia articles, as specified by the
Kyoto Free Translation Task (KFTT)
\footnote{\url{http://www.phontron.com/kftt}}.  The training,
development and test data of KFTT consist of 440k, 1166 and 1160 sentences
respectively.

We used GNMT \cite{wu2016google} as the implementation of the NMT system
in our experiments.  We generally followed the settings and training
procedure described in \cite{wu2016google}, however, we changed the node
and layer size of LSTM to be 512 and 6 respectively.

A word model is used as a baseline system. We compared to SentencePiece
(unigram language model) with and without
pre-tokenization. SentencePiece with pre-tokenization is essentially the
same as the common NMT configuration with subword-nmt. SentencePiece
without pre-tokenization directly trains the subword model from raw
sentences and does not use any external resources.  We used the Moses
tokenizer\footnote{\url{http://www.statmt.org/moses/}} and
KyTea\footnote{\url{http://www.phontron.com/kytea}} for English and Japanese
pre-tokenization respectively. The same tokenizers are applied to the
word model.

We used the case-sensitive BLEU score \cite{papineni2002bleu} as an
evaluation metric.  As the output sentences are not segmented in
Japanese, we segmented them with KyTea for before calculating BLEU scores.

\begin{table}[t]
      \renewcommand{\arraystretch}{0.9}
  \begin{center}
    \begin{tabular}[c]{c|l|l|l}
     \hline
     {\scriptsize Lang pair} & {\small setting (source/target)}  &
     {\small\shortstack{\# vocab.}} & BLEU \\
     \hline
     {\small ja$\rightarrow$en}
     & {\small Word model (baseline)}              & 80k/80k     & 28.24 \\
     & {\small SentencePiece}      & 8k {\scriptsize (shared)}& 29.55 \\
     & {\small SentencePiece w/ pre-tok.\!\!} & 8k {\scriptsize (shared)}& 29.85 \\
     & {\small Word/SentencePiece} & 80k/8k & 27.24 \\
     & {\small SentencePiece/Word} & 8k/80k & 29.14 \\
     \hline
     {\small en$\rightarrow$ja} & {\small Word model (baseline)} & 80k/80k     & 20.06 \\
     & {\small SentencePiece}      & 8k {\scriptsize (shared)} & 21.62 \\
     & {\small SentencePiece w/ pre-tok.\!\!} & 8k {\scriptsize (shared)}& 20.86 \\
     & {\small Word/SentencePiece} & 80k/8k & 21.41\\
     & {\small SentencePiece/Word} & 8k/80k & 19.94 \\
     \hline
    \end{tabular}
  \end{center}
  \vspace*{-4mm}
 \caption{Translation Results (BLEU(\%))}
 \label{result}
 \vspace*{-5mm}
\end{table}
Table \ref{result} shows the experimental results.  First,
as can be seen in the table, subword segmentations with SentencePiece
consitently improve the BLEU scores compared to the word model. This
result is consistent with previous work
\cite{sennrichneural}. Second, it can be seen that the pre-tokenization
is not always necessary to boost the BLEU scores. In Japanese to
English, the improvement is marginal and has no significant
difference. In English to Japanese, the BLEU score is degraded with
pre-tokenization.

We can find larger improvements in BLEU when 1) SentencePiece is
applied to Japanese, and 2) the target sentence is Japanese.  As
Japanese is a non-segmented language, pre-tokenization acts as a strong
constraint to determine the final vocabulary.  It can be considered that
the positive effects of unsupervised segmentation from raw input worked
effectively to find the domain-specific vocabulary in Japanese.

\subsection{Segmentation performance}
Table \ref{result2} summarizes the training and segmentation
performance of various configurations.

We can see that the training and segmentation speed of both
SentencePiece and subword-nmt is almost comparable on English data set
regardless of the choice of pre-tokenization. This is expected, as
English is a segmented language and the search space for the vocabulary
extraction is largely restricted.  On the other hand, SentencePiece
shows larger performance improvements when applying it to raw Japanese
data (w/o pre-tok). The segmentation speed of SentencePiece is about 380
times faster than that of subword-nmt in this setting. This result
strongly supports our claim that SentencePiece is fast enough to be
applied to raw data and the pre-tokenization is not always
necessary. Consequently, SentencePiece helps to build a purely
data-driven and language-independent system.
The segmentation speed of SentencePiece is around 21k and 74k
sentences/sec. in English and Japanese respectively, which is fast
enough to be executed on-the-fly.

\begin{table}[t]
 \renewcommand{\arraystretch}{0.9}
 \begin{center}
  \begin{tabular}[c]{l|c|c|r|r}
   \hline
   &           &      & \multicolumn{2}{c}{time (sec.)} \\
   \cline{4-5}
   {\small Task} & {\footnotesize Tool} & {\small Pre-tok.} & {\small Japanese}  &  {\small English} \\
   \hline
   {\small Train} & {\small subword-nmt}    & {\small yes}  & 56.9   & 54.1  \\
                  & {\small SentencePiece}  & {\small yes}  & 10.1   & 16.8   \\
                  & {\small subword-nmt}    & {\small no}   & 528.0  & 94.7  \\
                  & {\small SentencePiece}  & {\small no}   & 217.3  & 21.8  \\
   \hline
   {\small Seg.}  & {\small subword-nmt}    & {\small yes}   & 23.7   & 28.6   \\
                 & {\small SentencePiece}  & {\small yes}   & 8.2    & 20.3   \\
                 & {\small subword-nmt}    & {\small no}    & 216.2  & 36.1   \\
                 & {\small SentencePiece}  & {\small no}    & 5.9    & 20.3   \\
   \hline
   \multicolumn{3}{c|}{{\small Pre-tokenizaion}{\scriptsize\,\,KyTea(ja)/Moses(en)}} & 24.6   & 15.8    \\
   \hline
  \end{tabular}
  \vspace*{-4mm}
  \caption{Segmentation performance.  {\footnotesize KFTT corpus (440k
  sentences) is used for evaluation. Experiments are executed on
  Linux with Xeon 3.5Ghz processors. The size of vocabulary is
  16k. Moses and KyTea tokenizers are used for English and Japanese
  respectively. Note that we have to take the time of pre-tokenization
  into account to make a fair comparison with and without
  pre-tokenization.  Because subword-nmt is based on BPE, we used the
  BPE model in SentencePiece.  We found that BPE and unigram language
  models show almost comparable performance.}}  \label{result2}
  \vspace*{-6mm}
 \end{center}
\end{table}

\section{Conclusions}
In this paper, we introduced SentencePiece, an open-source subword
tokenizer and detokenizer designed for Neural-based text processing.
SentencePiece not only performs subword tokenization, but directly
converts the text into an id sequence, which helps to develop a purely
end-to-end system without replying on language specific resources.  The
model file of SentencePiece is designed to be self-contained to
guarantee perfect reproducibility of the normalization and subword
segmentation. We hope that SentencePiece will provide a stable and
reproducible text processing tool for production use and help the
research community to move to more language-agnostic and multilingual
architectures.

\bibliography{main} \bibliographystyle{acl_natbib_nourl}

\begin{thebibliography}{15}
\expandafter\ifx\csname natexlab\endcsname\relax\def\natexlab#1{#1}\fi

\bibitem[{Artetxe et~al.(2017)Artetxe, Labaka, Agirre, and Cho}]{mikel2017}
Mikel Artetxe, Gorka Labaka, Eneko Agirre, and Kyunghyun Cho. 2017.
\newblock Unsupervised neural machine translation.
\newblock \emph{arXive preprint arXiv:1710.11041}.

\bibitem[{Bahdanau et~al.(2014)Bahdanau, Cho, and Bengio}]{bahdanau2014neural}
Dzmitry Bahdanau, Kyunghyun Cho, and Yoshua Bengio. 2014.
\newblock Neural machine translation by jointly learning to align and
  translate.
\newblock \emph{arXiv preprint arXiv:1409.0473}.

\bibitem[{Denkowski and Neubig(2017)}]{denkowski2017stronger}
Michael Denkowski and Graham Neubig. 2017.
\newblock Stronger baselines for trustable results in neural machine
  translation.
\newblock \emph{Proc. of Workshop on Neural Machine Translation}.

\bibitem[{Johnson et~al.(2016)Johnson, Schuster et~al.}]{johnson2016google}
Melvin Johnson, Mike Schuster, et~al. 2016.
\newblock Google's multilingual neural machine translation system: enabling
  zero-shot translation.
\newblock \emph{arXiv preprint arXiv:1611.04558}.

\bibitem[{Kudo(2018)}]{kudo2018}
Taku Kudo. 2018.
\newblock Subword regularization: Improving neural network translation models
  with multiple subword candidates.
\newblock In \emph{Proc. of ACL}.

\bibitem[{Lample et~al.(2017)Lample, Denoyer, and Ranzato}]{guillaume2017}
Guillaume Lample, Ludovic Denoyer, and Marc'Aurelio Ranzato. 2017.
\newblock Unsupervised machine translation using monolingual corpora only.
\newblock \emph{arXive preprint arXiv:1711.00043}.

\bibitem[{Luong et~al.(2015)Luong, Pham, and Manning}]{luong2015effective}
Minh-Thang Luong, Hieu Pham, and Christopher~D Manning. 2015.
\newblock Effective approaches to attention-based neural machine translation.
\newblock In \emph{Proc of EMNLP}.

\bibitem[{Nakazawa et~al.(2017)Nakazawa, Higashiyama
  et~al.}]{nakazawa2017overview}
Toshiaki Nakazawa, Shohei Higashiyama, et~al. 2017.
\newblock Overview of the 4th workshop on asian translation.
\newblock In \emph{Proceedings of the 4th Workshop on Asian Translation
  (WAT2017)}.

\bibitem[{Papineni et~al.(2002)Papineni, Roukos, Ward, and
  Zhu}]{papineni2002bleu}
Kishore Papineni, Salim Roukos, Todd Ward, and Wei-Jing Zhu. 2002.
\newblock Bleu: a method for automatic evaluation of machine translation.
\newblock In \emph{Proc. of ACL}.

\bibitem[{Post(2018)}]{post2018}
Matt Post. 2018.
\newblock A call for clarity in reporting bleu scores.
\newblock \emph{arXiv preprint arXiv:1804.08771}.

\bibitem[{Rush et~al.(2015)Rush, Chopra, and Weston}]{rush2015neural}
Alexander~M Rush, Sumit Chopra, and Jason Weston. 2015.
\newblock A neural attention model for abstractive sentence summarization.
\newblock In \emph{Proc. of EMNLP}.

\bibitem[{Sennrich et~al.(2016)Sennrich, Haddow, and Birch}]{sennrichneural}
Rico Sennrich, Barry Haddow, and Alexandra Birch. 2016.
\newblock Neural machine translation of rare words with subword units.
\newblock In \emph{Proc. of ACL}.

\bibitem[{Vaswani et~al.(2017)Vaswani, Shazeer, Parmar, Uszkoreit, Jones,
  Gomez, Kaiser, and Polosukhin}]{ashish2017google}
Ashish Vaswani, Noam Shazeer, Niki Parmar, Jakob Uszkoreit, Llion Jones,
  Aidan~N. Gomez, Lukasz Kaiser, and Illia Polosukhin. 2017.
\newblock Attention is all you need.
\newblock \emph{arXive preprint arXiv:1706.03762}.

\bibitem[{Vinyals and Le(2015)}]{vinyals20152}
Oriol Vinyals and Quoc~V. Le. 2015.
\newblock A neural conversational model.
\newblock In \emph{ICML Deep Learning Workshop}.

\bibitem[{Wu et~al.(2016)Wu, Schuster et~al.}]{wu2016google}
Yonghui Wu, Mike Schuster, et~al. 2016.
\newblock Google's neural machine translation system: Bridging the gap between
  human and machine translation.
\newblock \emph{arXiv preprint arXiv:1609.08144}.

\end{thebibliography}

\end{document}